\documentclass[runningheads]{llncs}
\usepackage{graphicx}

\usepackage{tikz}
\usepackage{comment}
\usepackage{amsmath,amssymb} 
\usepackage{color}
\usepackage{marvosym}

\usepackage[accsupp]{axessibility}  

\begin{document}
\pagestyle{headings}
\mainmatter
\def\ECCVSubNumber{4972}  

\title{diffConv: Analyzing Irregular Point Clouds with an Irregular View} 

\titlerunning{diffConv: Analyzing Irregular Point Clouds with an Irregular View}
%
\author{Manxi Lin\textsuperscript{\Letter} \and
Aasa Feragen}
\authorrunning{Lin. et al.}
%
\institute{Technical University of Denmark, Kongens Lyngby, Denmark\\
\email{\{manli,afhar\}@dtu.dk}}

\maketitle

\begin{abstract}
Standard spatial convolutions assume input data with a regular neighborhood structure. Existing methods typically generalize convolution to the irregular point cloud domain by fixing a regular "view" through e.g.~a fixed neighborhood size, where the convolution kernel size remains the same for each point. However, since point clouds are not as structured as images, the fixed neighbor number gives an unfortunate inductive bias. We present a novel graph convolution named Difference Graph Convolution (diffConv), which does not rely on a regular view. diffConv operates on spatially-varying and density-dilated neighborhoods, which are further adapted by a learned masked attention mechanism. Experiments show that our model is very robust to the noise, obtaining state-of-the-art performance in 3D shape classification and scene understanding tasks, along with a faster inference speed. The code is publicly available at https://github.com/mmmmimic/diffConvNet.  
\end{abstract}

\section{Introduction}
The availability and affordability of 3D sensors are rapidly increasing. As a result, the point cloud --- one of the most widely-employed representations of 3D objects --- is used extensively in computer vision applications such as autonomous driving, medical robots, and virtual reality \cite{qi2017pointnet++,dometios2017real,blanc2020genuage}. The industrial demand has prompted a strong call for exploiting the underlying geometric information in points, fuelling deep learning models that rely on learning effective local point features~\cite{qi2017pointnet++,wang2019dynamic,wang2019graph,li2019deepgcns}.
\begin{figure}
 \centering
    \includegraphics[width=\linewidth]{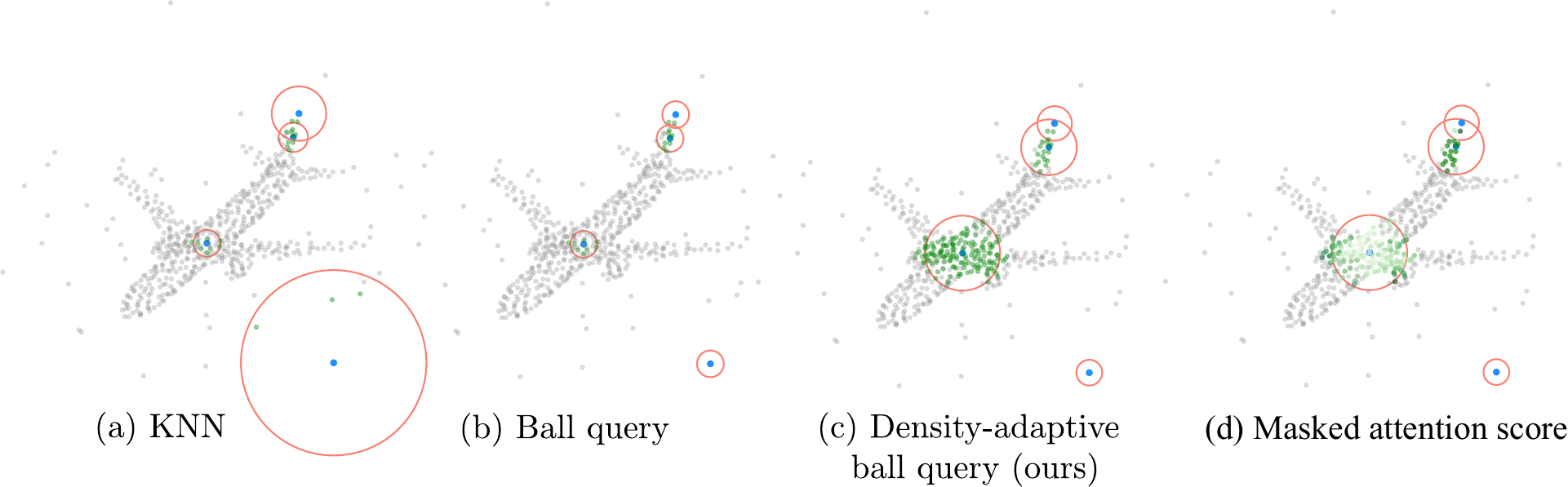}
  \caption{The blue dots refer to key points, green dots denote neighbors, orange circles indicate neighborhoods of key points. The color scale in (d) denotes the neighbor importance of the key point on the airplane body, given by masked attention score.}
  \label{fig: three groupings}
\end{figure}

Local feature learning schemes~\cite{qi2017pointnet++,xiang2021walk,wang2019dynamic} typically consist of two main steps, namely \emph{point grouping} and \emph{feature aggregation}. Point grouping gathers the neighbors of key points, to be encoded and fused in the subsequent feature aggregation. 

Unlike grid data, points are unordered and unstructured, posing great challenges to the processing; indeed, standard convolutional neural networks (CNNs) \cite{he2016deep,tan2019efficientnet} only work with grid data in a regular \emph{view}, or neighborhood. Point grouping and feature aggregation are often designed to imitate the regular view, or fixed neighborhood size and structure, that we know from convolutions on images. This imitation imposes inductive biases on the point cloud structure.

Fig.~\ref{fig: three groupings} illustrates problems associated with these inductive biases for the two most common regular point groupings, namely the $k$ nearest neighbor (KNN) grouping~\cite{fix1952discriminatory} and ball query~\cite{qi2017pointnet++}. KNN fixes its view to the $k$ nearest neighbors of each key point, and the resulting groupings are sensitive to noise and outliers. These are automatically grouped with far away points, sometimes belonging to the object of interest, strongly affecting their downstream features (Fig.~\ref{fig: three groupings}a). 

Ball query, conversely, constrains its view to a pre-defined constant radius ball. We see from (Fig.~\ref{fig: three groupings}b) that using the same radius, or view, at every location also causes problems: A small radius is needed to avoid grouping noise with far away points as we saw with KNN, but this means that the less dense parts of the airplane, such as its tail, easily become disconnected and mistaken for noise.

We thus see that the inductive bias imposed by regular views leads to problems. Can we process the unstructured point cloud without regular groupings? \textbf{In this paper, we analyze the irregular point cloud with an \emph{irregular view}.} We present a novel graph convolution operator, named Difference Graph Convolution (diffConv), which takes an irregular \emph{view} of the point cloud. In diffConv, we improve ball query with density-dilated neighborhoods where the radius for each point depends on its kernel density. This induces asymmetric neighborhood graphs, and in Fig.~\ref{fig: three groupings}c) this e.g.~helps the plane hold on to its tail. 

The dilated ball query still treats all the neighbors inside the ball equally. We further adopt masked attention to introduce an additional, task-specific \emph{learned} irregularity to the neighborhood. As illustrated in Fig~\ref{fig: three groupings}d, the points close to the key point get less attention than those further away, indicating that the points near the wings, where contextual differences appear, are paid more attention.   

\paragraph{In short, our key contributions are as follows:}
\begin{itemize}
    \item We propose diffConv, a local feature aggregator which does not rely on the regular view constraint.
    \item We present density-dilated ball query, which adapts its view to the local point cloud density, in particular alleviating the influence of noise and outliers. 
    \item We are the first to introduce masked attention to local point feature aggregation. In combination with Laplacian smoothing, it allows us to learn further irregularities, e.g.~focusing on contextual feature differences.
    \item We build a hierarchical learning architecture by stacking diffConv layers. Extensive experiments demonstrate that our model performs on par with state-of-the-art methods in noise-free settings, and outperforms state-of-the-art by far in the presence of noise.
\end{itemize}

\section{Background and Related Work}
Imitating the success of CNNs in local grid feature learning, most 3D point cloud models attempt to generalize CNNs to point clouds by fixing a regular view. 

\paragraph{Local point feature learning with projection-based methods.}
Early works~\cite{su2015multi,wang2018msnet,zhang20183d} convert the point cloud to regular 2D or 3D grids (voxels) and leverage the well-established standard image convolutions. Su et al.~\cite{su2015multi} projected the point cloud to multiple views and utilized 2D convolutions with max-pooling to describe the global features of points. MSNet~\cite{wang2018msnet} partitions the point cloud into multi-scale voxels, where 3D convolutions are applied to learn discriminative local features. These methods recast the point cloud to a regular view via the convolution-friendly grid representation. Nevertheless, they suffer from the expensive computation and inevitable loss of geometric detail due to discretization.

\paragraph{Point-based methods.} In contrast to the expensive projection or voxelization, point-based methods \cite{komarichev2019cnn,klokov2017escape,li2018so} process the point cloud directly and efficiently. The pioneering PointNet~\cite{qi2017pointnet} learns point features independently with multi-layer perceptrons (MLPs) and aggregates them with max-pooling. Since point-wise MLPs cannot capture local geometric structures, follow-up works obtain local point features via regular views from different types of point grouping.

\paragraph{Point grouping.} The KNN point grouping does not guarantee balanced density~\cite{liu2020closer} and adopts different region scales at different locations of the point cloud~\cite{qi2017pointnet++}. It thus comes with an inductive bias that makes it sensitive both to noise and outliers in point clouds, as well as to different local point densities, as also shown in Fig.~\ref{fig: three groupings}. Moreover, since KNN queries a relatively small neighborhood in the point-dense regions, the features encoded in subsequent aggregation might not contain enough semantic information to be discriminative. To keep the noise and outliers isolated, some works~\cite{qi2017pointnet++,liu2019relation} adopt ball query with an upper limit of $k$ points. Balls with less than $k$ points are filled by repeating the first-picked point. Neighborhoods with more than $k$ points are randomly subsampled. The upper limit $k$ prevents the neighborhood from having too many neighbors, but does also lead to information loss. Besides, the constant ball radius and limit $k$ are not adaptive to the uneven density of real-world point clouds.   

\paragraph{Local point feature learning with point-based methods.} PointNet++~\cite{qi2017pointnet++} suggests a hierarchical learning scheme for local point features, including point grouping and feature aggregation. Hierarchically, PointNet++ focuses on a set of subsampled \emph{key points}, which are represented by a neighborhood defined via point grouping according to pre-defined rules. Subsequently, point aggregation encodes and fuses the features of neighboring points to represent the latent shape of the point cloud structure. These point-based local feature learning methods regularize the input point cloud in the spatial or spectral domain. The point grouping in PointNet++ is done via ball query, and PointNet++ thus also inherits the corresponding inductive biases. Although the multi-scale and multi-resolution feature learning in PointNet++ can handle the uneven density to a degree, they are too expensive for application. 

Graph convolutional neural networks (GCNs)~\cite{kipfsemi,wu2019simplifying,defferrard2016convolutional,zhang2018end} generalize standard convolutions to graph data by considering convolutions in the spectral domain. Some methods~\cite{pan20183dti,te2018rgcnn,wang2018local} adopt GCNs for point aggregation. These methods represent the point cloud as an undirected graph for spectral analysis, where neighbors are constrained to a fixed radius ball. Thus, these methods obtain a regular view via spectral analysis, effectively using a form of ball query. It follows that the resulting algorithms also come with the inductive biases associated with ball query. Additionally, GCNs entail extra computations for signal transformation to the spectral domain. 

Alternative methods find other ways to define a $k$-size kernel for each key point, often based on KNN. In particular, DGCNN~\cite{wang2019dynamic} groups points by their $k$ nearest neighbors in feature space, and then aggregates the feature difference between points and their neighbors with max-pooling. In a similar way, CurveNet~\cite{xiang2021walk} takes guided walks and arguments the KNN neighborhood by hypothetical curves. Other works~\cite{wang2019graph,tian2021dnet,yang2020attpnet,xu2021paconv,zhao2021point} learn the importance weights of the KNN features to introduce some neighborhood irregularity, but are still confined to the fixed, and sometimes spatially limited, KNN view.  

Differing from the previous generalizations of CNNs, which all rely on taking a regular view of the point clouds in order to define convolutional operators, we suggest an irregular view of point cloud analysis in this paper. 

\section{Method}
We first revisit the general formulation of local feature learning in Section~\ref{sec: revisit}. Then we propose a flexible convolution operator, namely difference graph convolution in Section~\ref{sec: diffPool}. In Section~\ref{sec: DAND} and Section~\ref{sec: MA}, we enrich the basic diffConv with density-dilated ball query and masked attention. We present our network architectures for 3D shape classification and segmentation in supplements. 
\subsection{Revisit Local Point Feature Learning}
\label{sec: revisit}
Given a point cloud consisting of $N$ points $\mathcal{P}=\{p_i| i=1, 2, ..., N\}\in \mathbb{R}^{N \times 3}$, and a data matrix containing their feature vectors $X={[x_1, x_2, x_3, ..., x_N]}^T\in \mathbb{R}^{N \times d}$ for each point $p_i$, local feature extraction can be formulated as:
\begin{align}
g_i = \Lambda(h(p_i, p_j, x_i, x_j)|p_j \in \mathcal{N}(p_i))
\end{align}
where $g_i$ denotes the learned local feature vector, $\mathcal{N}(p_i)$ refers to the neighbors of $p_i$, $h(\cdot)$ is a function encoding both the coordinates $p_i$ of the $i^{th}$ point, and the coordinates $p_j$ of its neighbors, as well as their feature vectors $x_i$ and $x_j$, respectively. Moreover, $\Lambda$ denotes an aggregation function such as MAX or AVG.  

Most previous works focus on the design of $h(\cdot)$~\cite{wang2019dynamic,zhao2021point,zhou2021adaptive,wang2019graph}. Among them, one of the most widely applied methods is the edge convolution (edgeConv) appearing in DGCNN~\cite{wang2019dynamic}. In edgeConv, the neighborhood of a point is defined by its KNN in the feature space. The feature difference $x_i - x_j$ is used to indicate the pair-wise local relationship between the point and its neighbor. The relationships are concatenated with the original point features. The concatenated pair is processed by a multi-layer perceptron $l_\theta$ and summarized by a MAX aggregation. The whole process can be described as
\begin{align}
    g_i = M\!A\!X(l_\theta(x_i-x_j||x_i))
    \label{eq: edgeconv}
\end{align}
where $||$ refers to concatenation, and $p_j \in \mathcal{N}(p_i)$. In this paper, DGCNN is included as a baseline model. In edgeConv, KNN brings convenience to matrix-level operations, such as feature concatenation and linear transformation. One challenge tackled in this paper is the generalization of edgeConv to the irregular view, and in particular inclusion of feature differences into the convolution.    
\subsection{Difference Graph Convolution}
\label{sec: diffPool}
\begin{figure}[t]
    \centering
    \includegraphics[width=\linewidth]{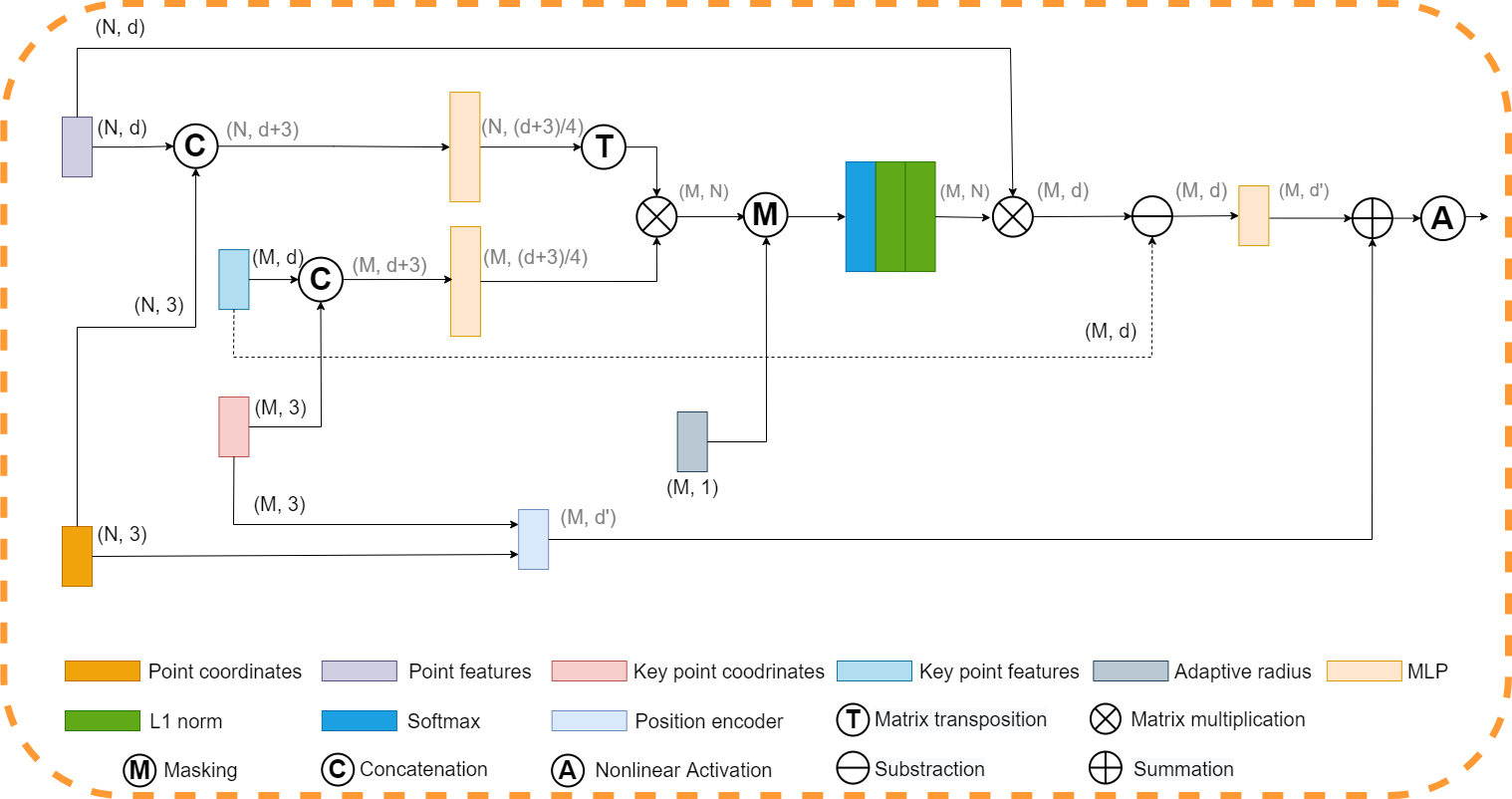}
    \caption{\textbf{Structure of our diffConv. }Specifically, masking denotes the conditional statements in Eq.~\ref{eq: ADJ}, which takes the adaptive radius $r_i$ from density-dilated ball query. In our implementation, $d_k$ was set to $\frac{(d+3)}{4}$. $M$ is the number of sampled key points. The dotted line is used to avoid overlapping with the solid lines.}
    \label{fig:diffConv}
\end{figure}

Inspired by GCNs~\cite{kipfsemi} and edgeConv~\cite{wang2019dynamic}, we propose the Difference Graph Convolution (diffConv). We present a basic version of diffConv in this subsection, with further improvements described in the following subsections.  Fig.~\ref{fig:diffConv} gives an overview of the complete diffConv, where the querying radius is pre-computed.   

By treating each point as a node, we represent the aforementioned point cloud $\mathcal{P}$ as a directed graph $\mathcal{G}=\{X, A\}$, where $A$ is the adjacency matrix, and $X$ refers to point feature vectors. In $\mathcal{G}$, each point is connected to its neighbors. The commonly-adopted neighbor-wise feature difference in edgeConv has been proven to be very effective in capturing the local structure of point clouds~\cite{li2019deepgcns,wang2019dynamic,xiang2021walk}. When processing graphs, the Laplacian matrix efficiently measures the difference between nodes and their neighbors. Borrowing from GCNs~\cite{kipfsemi}, we apply Laplacian smoothing on the point feature vector by
\begin{align}
    S = X - \hat{A}X
\label{eq: laplacian}
\end{align}
where ${S=[s_1, s_2, ..., s_N]}^T$ contains the updated feature vectors and $\hat{A}$ is the normalized adjacency matrix. Like edgeConv, our Difference Graph Convolution is an MLP $l_\theta$ on the combination of $S$ and the original point features $X$ 
\begin{align}
G = l_\theta (S || X)
\label{eq: diffConv}
\end{align}
where $G={[g_1, g_2, ..., g_N]}^T$ includes the learned feature vectors. Here, $X$ is concatenated to present the global shape information of the point cloud. 

\paragraph{Relationship to edgeConv.}We argue that our diffConv is a variant of edgeConv in a more flexible neighborhood. When applying diffConv on KNN grouping, it only differs in the sequence of nonlinear activation and aggregation from edgeConv. As a demonstration, we consider a basic diffConv, with a simple binary adjacency matrix $A$ whose entries indicate connections between points. When the neighborhood is defined by ball query (without an upper limit on the number of neighbors), the connection between $p_i$ and $p_j$, $A_{ij}$, is given by 
\begin{align}
    A_{ij} = \begin{cases}
    1 & \text{if } {\lVert p_i - p_j\rVert}_2 < r\\
    0 & \text{otherwise}
    \end{cases} 
    \label{eq: constr}
\end{align}
where $r$ is the priory-given ball query search radius. An arithmetic mean is leveraged to normalize the adjacency matrix. According to Eq.~\ref{eq: laplacian}, for $p_j \in \mathcal{N}(p_i)$, the smoothed feature vector $s_i$ calculated by
\begin{align}
    s_i = x_i - A\!V\!G(x_j) =  A\!V\!G(x_i-x_j)
\label{eq: laplacian_element}
\end{align}
is the average of the neighbor-wise feature difference from Eq.~\ref{eq: edgeconv}. In line with Eq.~\ref{eq: diffConv}, diffConv for point $p_i$ becomes
$
    g_i = 
    l_\theta \left(A\!V\!G(x_i-x_j)||x_i\right) .  
$
When $l_\theta$ is linear, diffConv can be further decribed by
$
    g_i = A\!V\!G(l_\theta(x_i-x_j||x_i)) . 
$
Compared with Eq.~\ref{eq: edgeconv}, in the case of linear activation, the only difference between diffConv and edgeConv is in the aggregation function. When $l_\theta$ is nonlinear, the feature difference is activated before aggregation, while diffConv operates in reverse order. In contrast to edgeConv, our diffConv is more flexible with respect to the definition of neighborhood - the number of neighbors is no longer fixed.

\subsection{Density-dilated Ball Query}
\label{sec: DAND}
In Eq.~\ref{eq: constr}, the point cloud is constructed as an undirected graph, where $A_{ij}=A_{ji}$ for point pair $p_i$ and $p_j$. The directed relations between points and their neighbors are therefore neglected, as illustrated in  Fig.~\ref{fig: three groupings}b). We relax this to a density-dilated ball query where the search radius for each point is expanded according to its kernel density.
\begin{figure}[t]
    \centering
    \includegraphics[width=\linewidth]{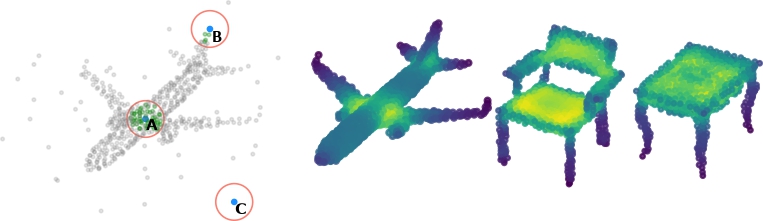}
    \caption{\textbf{Left:} Ball query in a point cloud with noise. The blue circles refer to key points, green circles denote neighbors, orange circles refer to neighborhoods, and gray circles denote other points in the point cloud. \textbf{Right:} Visualizations of the kernel density on some objects from ModelNet40~\cite{modelnet}. Brighter color indicates higher density. }
    \label{fig: density}
\end{figure}

Point density, defined as the number of neighbors of each point, reflects the spatial distribution of points. Fig.~\ref{fig: density} demonstrates a point cloud with noise. Note that under a fixed-radius ball query, contour points of the 3D object (e.g.~point $B$) have a lower density than the points from the flat area (e.g.~point $A$). Noise generates extreme cases (e.g.~point $C$), which can be isolated from the object. 

As illustrated in Fig.~\ref{fig: three groupings}, in KNN, points with low density, such as noise, get a spatially larger neighborhood than others, grouping them with unrelated features far away. The adversarial effect of noise points on estimated object properties is thus negatively correlated with their density, and assigning smaller neighborhoods to low-density points therefore contributes to resisting noise. 

Xiang et al. \cite{xiang2021walk} claim that boundary points provide different feature differences from neighbors while points from flat areas usually have similar and unrecognizable local feature differences. To enlarge the variance of feature difference, points with a higher density (points from flat areas) thus ought to have larger local regions.  
The KNN grouping queries small neighborhoods for points with high density. Our ball query enlarges the neighborhood to include more contextually different neighbors, giving access to more diverse information. 

Instead of counting neighbors, which is computationally expensive, we  describe the spatial distribution of points using kernel density estimation~\cite{turlach1993bandwidth}. For each point $p_i$, its kernel density $d_i$ is estimated by a Gaussian kernel
\begin{align}
    d_i = \frac{1}{Nh}\sum_{j=1}^{N}\frac{1}{\sqrt{2\pi}}e^{-\frac{{\Vert p_i - p_j\Vert}_2^2}{2h^2}}  ,
\end{align}
where $h$ is a parameter controlling the bandwidth. The density $d_i$ indicates the probability of point $p_i$ to be located in flat areas of the object, i.e., the degree of dilation. Fig.~\ref{fig: density} shows the kernel density of various objects, where points from flat areas have a higher density. This is in accordance with our former analysis. 

Based on this density, we softly dilate the searching radius from Eq.~\ref{eq: constr} to
\begin{align}
    r_i = \sqrt{r^2 (1+\hat{d}_i)} ,
\label{eq: DAND}
\end{align}
where the square and root operations are applied to slow down the dilation speed, $\hat{d_i}$ denotes the normalized kernel density within 0 and 1 by dividing the largest density. The dilated search radius varies from point to point, resulting in a directed graph. Since the search radius is enlarged to different degrees, the message propagation on the graph is boosted, accelerating the long-range information flow. Finally, our method is intuitively robust to noise, which is also demonstrated by our experiments in Section~\ref{sec: performance analysis}.  

\subsection{Masked Attention} 
\label{sec: MA}
The dilated ball query still treats all the neighbors inside the ball equally. We further introduce an additional, task-specific learned irregularity to the neighborhood by attention mechanism \cite{gehring2016convolutional,vaswani2017attention}. Different from other attention-based methods \cite{zhao2021point,wang2019graph}, we employ the masking mechanism in self-attention \cite{vaswani2017attention} to handle the no-longer-fixed neighborhood. We call this learning scheme masked attention. In a transformer built by self-attention \cite{vaswani2017attention} \cite{velikovi2017graph}, there are two kinds of masks: the padding mask in the encoder is exploited to address sequences with different lengths and the sequence mask in the decoder is used to prevent ground truth leakage in sequence prediction tasks. In contrast, we employ a neighborhood mask in our masked attention to learn the local features of points. It works in the encoder while playing a different role than the padding mask. It includes three steps, the calculation of edge weight for each pair of connected points on the graph, local normalization, and balanced renormalization.

First, we revise the adjacency matrix $A$ in Eq.~\ref{eq: constr} to
\begin{align}
    A_{ij} = \begin{cases}
    l_\phi(x_i||p_i){l_\psi(x_j||p_j)}^T & \text{if } {\lVert p_i - p_j\rVert}_2 < r_i\\
    -\infty & \text{otherwise} ,
    \end{cases}
\label{eq: ADJ}
\end{align}
where $l_\phi$ and $l_\psi$ are two MLPs mapping the input to a given dimension $d_k$. The MLPs are employed to learn the latent relationship and alignment between $p_i$ and $p_j$ in the feature space. The MLPs encode both point features and coordinates, as we expect them to recognize both the semantic and geometric information. In the implementation, $-\infty$ is approximated by $-{10}^{9}$, which is orders of magnitude smaller than the computed attention score. 

The adjacency matrix is then normalized by softmax
\begin{align}
    \tilde{A}_{i,j} = \frac{e^{A_{i,j}}}{\sum_{k}^{N}e^{A_{i,k}}} .
\label{eq: masked_norm}
\end{align}
Since $e^{-{10}^{9}} \approx 0$, points outside the neighborhood are automatically masked. Hence, Eq.~\ref{eq: masked_norm} becomes the softmax normalization in neighborhoods. With the masking mechanism, we normalize the attention score locally with only matrix-wise operations instead of expensive iterations. In each local region, neighbor features are aggregated dynamically and selectively. 

In self-attention~\cite{vaswani2017attention}, the dot product in Eq.~\ref{eq: ADJ} is scaled by $\sqrt{d_k}$ before the softmax normalization, in order to prevent the product from getting extreme values as $d_k$ increases. Here, we refine the scaling by a balanced renormalization of the attention scores $\tilde{A}$ from Eq.~\ref{eq: masked_norm}. 
We apply the $l_1$-norm twice on the square root of the former attention score in different dimensions by
\begin{align}
    \overline{A}_{ij} = \frac{\sqrt{\tilde{A}_{ij}}}{\sum_{k}^{N}\sqrt{\tilde{A}_{kj}}}
\quad \textrm{ and } \quad 
    \hat{A}_{ij} = \frac{\overline{A}_{ij}}{\sum_k^N\overline{A}_{ik}} .
    \label{eq: BR}
\end{align} 
Here, the square root reduces the variation in attention scores, to prevent overly attention on a few points. Considering the matrix product in Eq.~\ref{eq: laplacian}, $l_1$-norm is to balance the contribution of feature channels and neighbor points respectively. The experimental results in supplemental materials illustrate the effectiveness of the proposed renormalization strategy. 

\paragraph{Position encoding.}
Local spatial information is defined as the coordinate difference between points and their neighbors. Many prior works have demonstrated the importance of local spatial information in point cloud analysis~\cite{wang2019graph,liu2019relation,zhou2021adaptive}. We employ the Local Point-Feature Aggregation block~\cite{xiang2021walk} as the position encoder in diffConv. The encoded coordinates are fused with point features by
 \begin{align}
    G = \sigma(l_\theta (S || X) + l_\pi(P)) ,
\label{PE}
 \end{align}
where $\sigma$ is a nonlinear activation function, $P={[p_1, p_2, p_3, ..., p_N]}^T\in \mathbb{R}^{N \times 3}$ contains the 3D coordinates of points in $\mathcal{P}$, and $l_\pi$ is the position encoder working in the dilated neighborhoods and mapping the positional difference to the same dimension as the updated point features from $l_\theta$.       

\section{Experiments}
We demonstrate our model performance in 3D shape classification, object shape part segmentation, and real-scene semantic segmentation tasks. Due to the limited space, we describe the training settings as well as the corresponding ablation studies on the hyperparameters in supplements.

\setlength{\tabcolsep}{4pt}
\begin{table}[t]
\begin{center}
\caption{3D shape classification results on ModelNet40 dataset with official split as well as ModelNet40-C, and part segmentation results on ShapeNet Part dataset. Here, 'xyz' stands for coordinates of points; 'nor' stands for normals of points; '\#points' means the number of input points in classification; 'OA' stands for the overall accuracy over instances on ModelNet40; and 'MA' stands for the mean accuracy within each category on ModelNet40; 'CER' denotes the corruption error rate on ModelNet40-C;  'mIOU' denotes the instance average Inter-over-Union on ShapeNetPart. Not all papers report MA and CER; these are recorded as '-' in the table.}
\label{tab:officialsplit}
\begin{tabular}{lllllll}
\hline\noalign{\smallskip}
Methods & Input & \#points & OA(\%) & MA(\%) & CER(\%) & mIOU (\%)\\
\noalign{\smallskip}
\hline
\noalign{\smallskip}
A-CNN\cite{komarichev2019cnn} & xyz+nor & $1$k & $92.6$ & - & - & $86.1$\\
Kd-Net\cite{klokov2017escape} & xyz & $32$k & $91.8$ & $88.5$ & - & $82.3$\\
SO-Net\cite{li2018so} & xyz+nor & $5$k & $93.4$ & $\bf90.8$ & - & $84.6$\\
\hline
\noalign{\smallskip}
PointNet\cite{qi2017pointnet} & xyz & 1k & $89.2$ & - & $28.3$ & $83.7$\\
PointNet++\cite{qi2017pointnet++} & xyz & $1$k & $90.7$ & - & $23.6$ & $85.1$\\
DGCNN\cite{wang2019dynamic} & xyz & $1$k & $92.9$ & $90.2$ & $25.9$ & $85.1$\\
CurveNet\cite{xiang2021walk} & xyz & $1$k & $\bf93.8$ & - & $22.7$ & $\bf86.6$\\
RS-CNN\cite{liu2019relation} & xyz & $1$k & $92.9$ & - & $26.2$ & $86.2$\\    
Spec-GCN~\cite{wang2018local} & xyz & $1$k & $91.5$ & - & - & $85.4$\\
PointASNL\cite{yan2020pointasnl} & xyz & $1$k & $92.9$ & - & - & $86.1$\\    
PCT\cite{guo2021pct} & xyz & $1$k & $93.2$ & - & $25.5$ & $86.4$\\
GDANet\cite{xu2020learning} & xyz & $1$k & $93.4$ & - & $25.6$ & $86.5$\\
\hline
\noalign{\smallskip}
Ours & xyz & $1$k & $93.6$ & $90.6$ & $\bf21.4$ & $85.7$\\
\hline
\end{tabular}
\end{center}
\end{table}
\setlength{\tabcolsep}{1.4pt}

\subsection{Point Cloud Classification}
\paragraph{ModelNet40 with the official split.} 
We conducted experiments performing object shape classification on the ModelNet40 dataset~\cite{modelnet}, which contains 12,311 synthetic meshed CAD models from 40 different categories. The official split includes 9,843 models for training and 2,468 models for testing. Following most publicly available implementations~\cite{wang2019dynamic,xiang2021walk,qi2017pointnet++}, our validation on the official split was made by reporting the best-observed test performance over all epochs. Similar to previous works, we represented each CAD model by 1,024 points uniformly sampled from mesh faces with $(x,y,z)$ coordinates.

We measured both the overall accuracy (OA), treating every point cloud with equal weight, and the mean accuracy (MA), where the mean was computed from the category-specific accuracies. Note that the MA up-weights the importance of small categories, giving them equal weight as the larger categories.

Our results can be found in Table~\ref{tab:officialsplit}. All results for other methods are taken from the cited papers. Note that RS-CNN, CurveNet, and GDANet adopt a voting strategy to improve the model performance. We report their classification accuracy without this trick for a fair comparison. Our model achieves the second-best overall accuracy at $93.6\%$. Even compared to SO-Net, which takes $5k$ points with normal as input, our model has a better performance in OA. 

\paragraph{ModelNet40 with resplits for assessing robustness. } 
In addition to benchmarking on the official split, we also performed repeated experiments on resampled dataset splits, to obtain an evaluation that is robust to over-fitting to any specific data split. Here, we selected parameters in a more traditional fashion with the help of an independent validation set. To this end, we randomly re-split the dataset into 8,617/1,847/1,847 point clouds for training/validation/testing, respectively, and picked the model with the best validation performance for evaluation on the test. This was repeated 10 times, reporting the mean and standard deviation of the corresponding performance measures. The baseline models were trained with settings from their paper.

\setlength{\tabcolsep}{4pt}
\begin{table}[t]
\begin{center}
  \caption{Classification results on ModelNet40 with resplits and ScanObjectNN. OBJ\_ONLY and OBJ\_BG denote the overall accuracy in the corresponding variant. }
  \label{tab:resplit}
\begin{tabular}{llllll}
\hline\noalign{\smallskip}
    Methods & OA(\%) & MA(\%) & OBJ\_ONLY & OBJ\_BG \\
\noalign{\smallskip}
\hline
\noalign{\smallskip}
    PointNet\cite{qi2017pointnet} & $90.19\pm0.58$ & $84.55\pm0.81$ & $79.2$ & $73.3$  \\
    DGCNN\cite{wang2019dynamic} & $92.89\pm0.43$ & $89.62\pm0.83$ & $86.2$ & $82.8$  \\
    CurveNet\cite{xiang2021walk} & $92.86\pm0.49$ & $89.51\pm0.81$ & $84.3$ & $84.4$ \\
\hline
\noalign{\smallskip}
    Ours & $\bf93.15 \pm \bf0.34$ & $\bf89.86 \pm \bf0.56$ & $\bf86.6$ & $\bf84.9$\\
\hline
\end{tabular}
\end{center}
\end{table}
\setlength{\tabcolsep}{1.4pt}

We compare our model performance with PointNet, DGCNN, and CurveNet, where the latter has the highest performance on the official split. According to Table~\ref{tab:resplit}, our method outperforms all the other models in both overall and class-mean accuracy, showing a better generalization capability on the resampled data splits than other methods. Moreover, our model has the smallest standard deviation in both OA and MA, indicating a more stable performance. This supports our hypothesis that diffConv's irregular view gives discriminative local features.

\paragraph{ModelNet40-C with corruptions for assessing robustness. }
To further assess the corruption robustness, we evaluated the model trained on ModelNet40 with the official split on the ModelNet40-C dataset~\cite{sun2022benchmarking}, which imposes 15 common and realistic corruptions with 5 severity levels on the original ModelNet40 point clouds. We used the corruption error rate (CER) as the evaluation metric. 

The corruption performance of different models is shown in Table~\ref{tab:officialsplit}. All results for other methods are taken from~\cite{sun2022benchmarking}. Our model outperforms all existing methods with at least a 1.3\% difference, showing strong robustness to corruption. 

\paragraph{ScanObjectNN.}
Besides the synthetic datasets, we also validated our model on the ScanObjectNN dataset~\cite{uy-scanobjectnn-iccv19}, with 2,902 scanned real-world objects from 15 categories, with 1,024 points each. Compared to ModelNet40, ScanObjectNN is more challenging due to the presence of noise, backgrounds, and occlusion. 80\% of the objects are split into the official training set and the remaining 20\% are used for testing. We conducted experiments on objects with and without background respectively (the OBJ\_BG and OBJ\_ONLY variant of the dataset). 

Table~\ref{tab:resplit} summarizes the performance of our model and the baselines. The results of PointNet and DGCNN are taken from~\cite{uy-scanobjectnn-iccv19}, while CurveNet was retrained as it is not shown in~\cite{uy-scanobjectnn-iccv19}. Our model gets the highest OA on both two variants, indicating that it is practical in addressing real-world problems.

\begin{figure*}[t]
    \centering
    \includegraphics[width=0.8\linewidth]{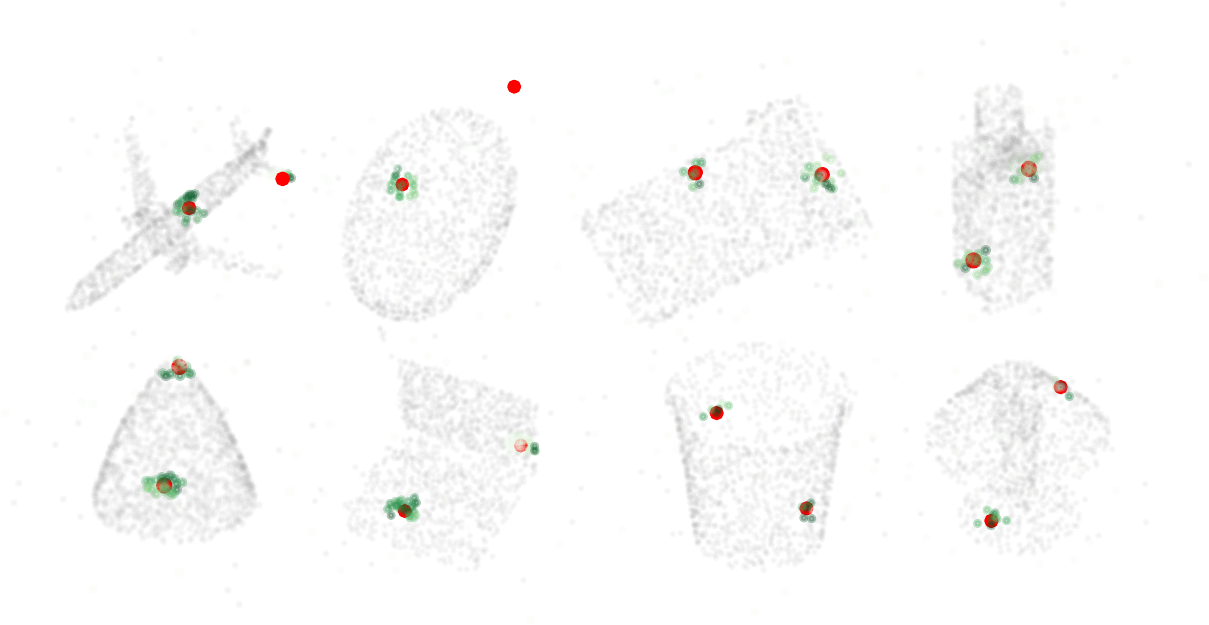}
    \caption{Example masked attention maps for ModelNet40-C. Red dots are picked key points; green dots denote neighbors; the color scale denotes the masked attention score. Zoom in for details. }
    \label{fig: att}
\end{figure*}

\paragraph{Visualizations of masked attention. }
Our model achieves state-of-the-art performance and shows strong robustness in 3D object classification. Fig.~\ref{fig: att} illustrates some examples of the attention maps from the second diffConv layer in our classification network on ModelNet40. We randomly picked two key points from 8 different objects of the ModelNet40-C benchmark. As shown in the figure, the noise points are isolated and the flat-area points have a larger neighborhood than the boundary points. Besides, neighbors with larger contextual differences to the query point, e.g., the edges of the airplane and the laptop, usually get more attention. More visualizations are presented in the supplements.  

\subsection{3D Semantic Segmentation}
We performed the scene-level understanding task on Toronto3D~\cite{tan2020toronto}. Compared with the indoor datasets, the Toronto3D dataset, collected by LiDAR in outdoor scenarios, contains more noise and requires more from the model's robustness. Toronto3D contains 8 different classes and 78.3M points. Following TGNet~\cite{li2019tgnet}, we divided the scene into 5mx5m blocks, and sampled 2,048 points from each block. There are 956 samples for training and 289 for testing. The mean intersection over union over categories(mean IoU) and IoU for each category are shown in Tab.~\ref{tab:toronto}, where the results of other methods are taken from~\cite{tan2020toronto}. Our model outperforms all other methods in mean IOU, and in many categories. Specifically, our model surpasses other methods significantly in recognizing small objects, such as road marks and poles. This demonstrates the effectiveness of our method on scene-level understanding tasks. 

\begin{table}[h]
\centering
\caption{Segmentation results on Toronto3D. Here, mean IoU denotes the category average Inter-over-Union in Toronto3D. }
\resizebox{\linewidth}{!}{
\begin{tabular}{llllllllll}
\hline\noalign{\smallskip}
Methods & mean IoU & Road & Rd. mrk. & Natural & Building & Util. line & Pole & Car & Fence\\
\hline
PointNet++~\cite{qi2017pointnet++} & 56.55 & \bf91.44 & 7.59 & 89.80 & 74.00 & 68.60 & 59.53 & 52.97 & 7.54\\
PointNet++(MSG)~\cite{qi2017pointnet++} & 53.12 & 90.67 & 0.00 & 86.68 & 75.78 & 56.20 & 60.89 & 44.51 & 10.19\\
DGCNN~\cite{wang2019dynamic} & 49.60 & 90.63 & 0.44 & 81.25 & 63.95 & 47.05 & 56.86 & 49.26 & 7.32\\
KPFCNN~\cite{thomas2019kpconv} & 60.30 & 90.20 & 0.00 & 86.79 & \bf86.83 & \bf81.08 & 73.06 & 42.85 & 21.57\\
MS-PCNN~\cite{ma2019multi} & 58.01 & 91.22 & 3.50 & 90.48 & 77.30 & 62.30 & 68.54 & 52.63 & 17.12\\
TG-Net~\cite{li2019tgnet} & 58.34 & 91.39 & 10.62 & 91.02 & 76.93 & 68.27 & 66.25 & 54.10 & 8.16\\
MS-TGNet~\cite{tan2020toronto} & 60.96 & 90.89 & 18.78 & \bf92.18 & 80.62 & 69.36 & 71.22 & 51.05 & 13.59\\
\hline\noalign{\smallskip}
Ours & \bf76.73 & 83.31 & \bf51.06 & 69.04 & 79.55 & 80.48 & \bf84.41 & \bf76.19 & \bf89.83\\
\hline
\end{tabular}
}
\label{tab:toronto}
\end{table}

\subsection{Shape Part Segmentation}
We conducted the 3D part segmentation task on the ShapeNetPart dataset~\cite{yi2016scalable}, which consists of 16,881 CAD models from 16 categories with up to 50 different parts. Each CAD model is represented by 2,048 uniformly-sampled points. We followed~\cite{wang2019graph} and trained our model on 12,137 instances with the rest for testing. All 2,048 points in each point cloud were used. 

\begin{figure*}[b]
    \centering
    \includegraphics[width=0.8\linewidth]{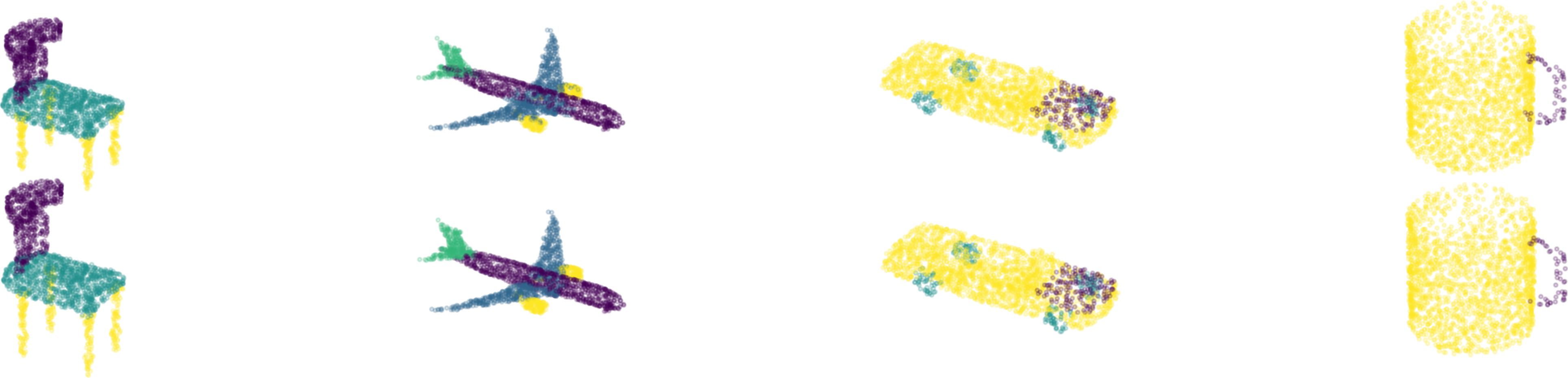}
    \caption{Examples of ShapeNetPart segmentation. Ground truth \textbf{(top)}; ours \textbf{(bottom)}.}
    \label{fig: seg_result}
\end{figure*}

We report the instance average Inter-over-Union (mIOU) in Table~\ref{tab:officialsplit}. The results for other methods are taken from the cited papers. Fig.~\ref{fig: seg_result} shows our model predictions, which are very close to the ground truth.

\subsection{Performance analysis}
\label{sec: performance analysis}

\begin{figure}[t]
    \includegraphics[width=\linewidth]{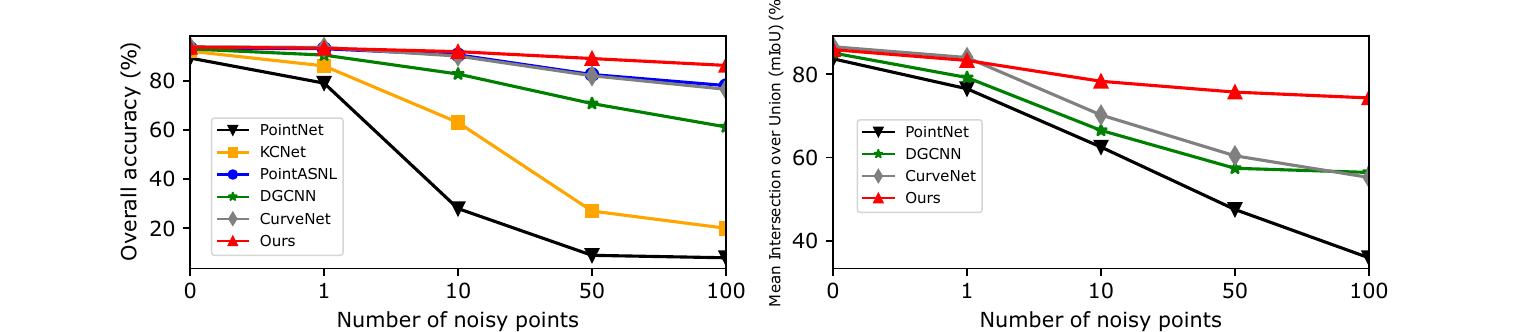}
  \caption{Results of robustness study on ModelNet40 (left) and ShapeNetPart (right). }
  \label{fig:noise}
\end{figure}

\paragraph{Robustness study.}
According to our analysis, our method is robust to noise, especially in contrast to KNN-based models. To further verify the robustness of our model, like KCNet~\cite{shen2018mining}, we evaluated the model on noisy points in both classification and part segmentation tasks. A certain number of randomly picked points were replaced with uniform noise ranging from -1 to 1. In segmentation, these noise points were labeled the same as the nearest original point. Fig.~\ref{fig:noise} demonstrates the results. We compared our model with the baseline methods used in previous experiments. In classification, we also report the performance of KCNet and PointASNL, with the results taken from papers. As presented in the figure, our model shows clear robustness to noise. Even with 100 noise points, our method has an OA of 86.2\% and an mIOU of 74.3\%. 

\paragraph{Computational resources}
We compare the corrupted error rate on ModelNet40-C, inference time (IT), floating-point operations (FLOPs) required, parameter number (Params), and memory cost against baselines in Table~\ref{tab:crr}. FLOPs and Params are estimated by PyTorch-OpCounter\footnote{https://github.com/Lyken17/pytorch-OpCounter}. The inference time per batch is recorded with a test batch size of 16. We conducted the experiments on a GTX 1080 Ti GPU. For a fair comparison in inference time and memory cost, each model is implemented in Python without CUDA programming. The table illustrates the efficiency of our method, which is twice as fast as CurveNet. 

\setlength{\tabcolsep}{4pt}
\begin{table}[h]
\begin{center}
  \caption{Computational resource requirements.}
  \label{tab:crr}
\begin{tabular}{lllllll}
\hline\noalign{\smallskip}
     Methods & CER(\%) & IT (ms) & FLOPs & Params & Memory\\
\noalign{\smallskip}
\hline
\noalign{\smallskip}
    PointNet \cite{qi2017pointnet} & 28.3 & \textbf{19} & 0.45G & 3.47M & \textbf{2061M}\\
    DGCNN \cite{wang2019dynamic} & 25.9 & 88 & 2.43G & 1.81M & 8067M\\
    CurveNet \cite{xiang2021walk} & 22.7 & 190 & 0.66G & 2.14M & 4309M\\ 
\hline
\noalign{\smallskip}
    Ours & 21.4 & 78 & \textbf{0.28G} & \textbf{1.31M} & 3971M\\
\hline
\end{tabular}
\end{center}
\end{table}
\setlength{\tabcolsep}{1.4pt}

\section{Conclusion}
We have reviewed contemporary local point feature learning, which heavily relies on a regular inductive bias on point structure. We question the necessity of this inductive bias, and propose diffConv, which is neither a generalization of CNNs nor spectral GCNs. Instead, equipped with Laplacian smoothing, density-dilated ball query and masked attention, diffConv analyzes point clouds through an \emph{irregular view}. Through extensive experiments we show that diffConv achieves state-of-the-art performance in 3D object classification and scene understanding, is far more robust to noise than other methods, with a faster inference speed.

\section*{Acknowledgements}
This work was supported by the DIREC project EXPLAIN-ME, the Novo Nordisk Foundation through the Center for Basic Machine Learning Research in Life Science (NNF20OC0062606), and the Pioneer Centre for AI, DNRF grant nr P1.

\bibliographystyle{splncs04}
\bibliography{egbib}
\appendix

\section{Network Architecture}
\begin{figure*}[h]
    \centering
    \includegraphics[width=\linewidth]{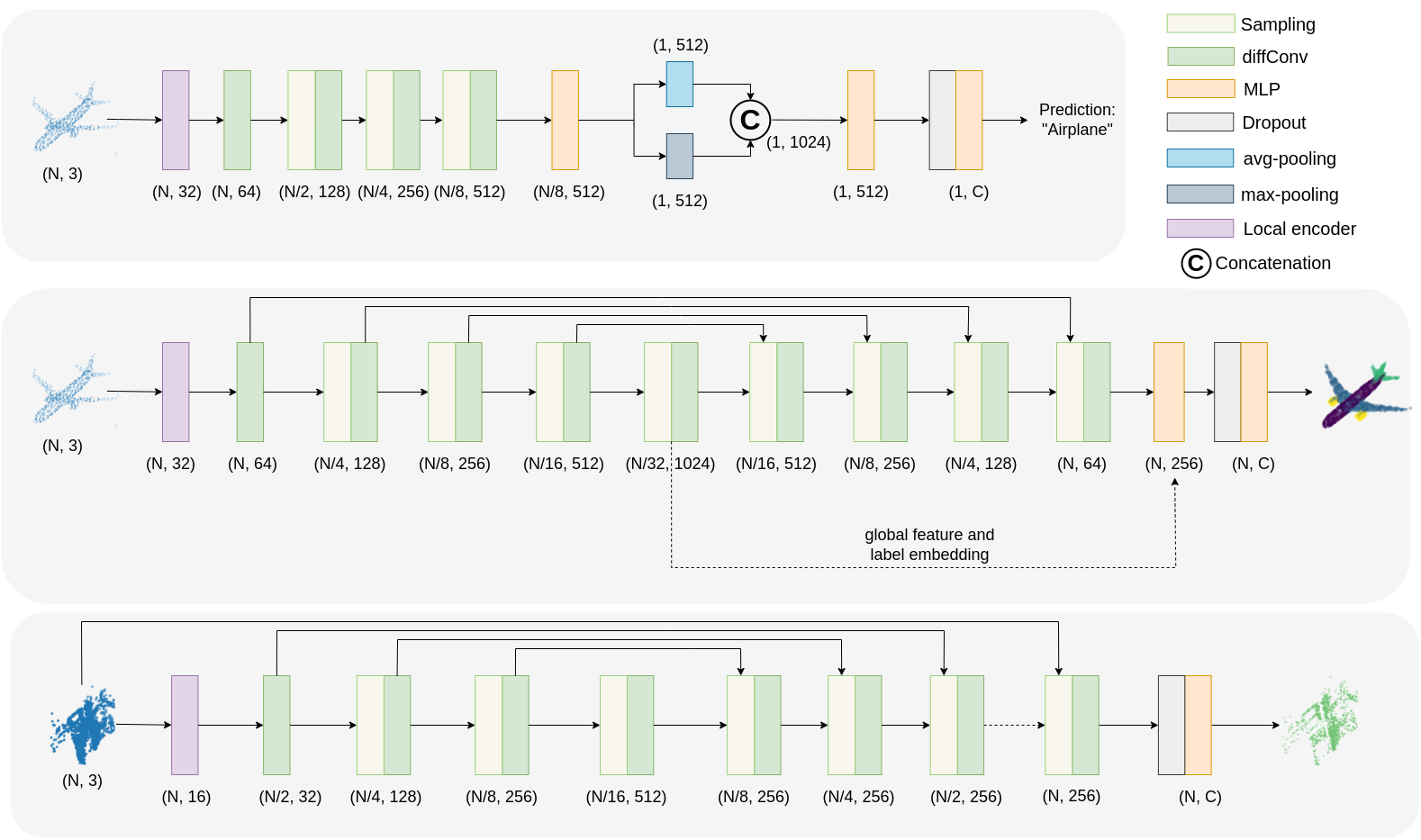}
    \caption{Overview network architectures for classification \textbf{(top)} 3D shape part segmentation \textbf{(middle)}, and scene semantic segmentation \textbf{(bottom)}. Specifically, in the encoders, 'Sampling' denotes random downsampling, while in the decoders, it denotes feature interpolation (point upsampling).}
    \label{fig: network}
\end{figure*}

Fig.~\ref{fig: network} shows our network for different tasks, where diffConv is performed hierarchically to capture multi-scale point features and avoid redundant computation.

Specifically, input point coordinates are initially encoded to a higher dimension by a local feature encoder. For classification, the encoder is an MLP, for segmentation, the encoder is a point abstraction module~\cite{qi2017pointnet++}. Then, point features are grouped and aggregated through multiple diffConvs. In contrast to the widely-adopted furthest point sampling~\cite{qi2017pointnet++}, during encoding, we select key points by random sampling, which has recently been proved efficient and effective~\cite{hu2020randla}. For classification, we follow the global aggregation scheme of DGCNN \cite{wang2019dynamic}, where the learned local features are pooled by a max-pooling and an average-pooling (avg-pooling) respectively. The pooled features are concatenated and processed by MLPs. For segmentation, we use the same attention U-Net style decoder architecture with CurveNet. In the 3D shape segmentation task, we fuse the global feature and the label embedding of object shape category with the learned features, following DGCNN~\cite{wang2019dynamic}.     

\section{Training Details}
\paragraph{Settings. }For all experiments, the squared initial searching radius $r^2$ was set to 0.005, and increased to the inverse of the sampling rate times after each sampling; the kernel density bandwidth $h$ was set to 0.1. Gaussian error linear units~\cite{hendrycks2016gaussian} were used as nonlinear activation. In 3D object classification and part segmentation, we trained the models by optimizing cross-entropy loss with label smoothing, using SGD with a learning rate of 0.1 and a momentum of 0.9, using batch size 32 in training and 16 in testing. In the scene segmentation task, the optimizer was AdamW with a learning rate of 0.001 and the training batch size was 16. The learning rate was reduced by cosine annealing to 0.001. The random seed was fixed at 42 in all experiments to enhance reproducibility. The dropout rate was set to 0.5, and we applied random scaling within $[\frac{2}{3}, \frac{3}{2}]$, random translation within $[-0.2, 0.2]$ and shuffling as augmentation. In scene segmentation, we also jittered the points within the range of $\pm0.01$ during training. We trained the model for 2,000 epochs in all the tasks. 

\begin{figure}[h]
  \centering
  \includegraphics[width=\linewidth]{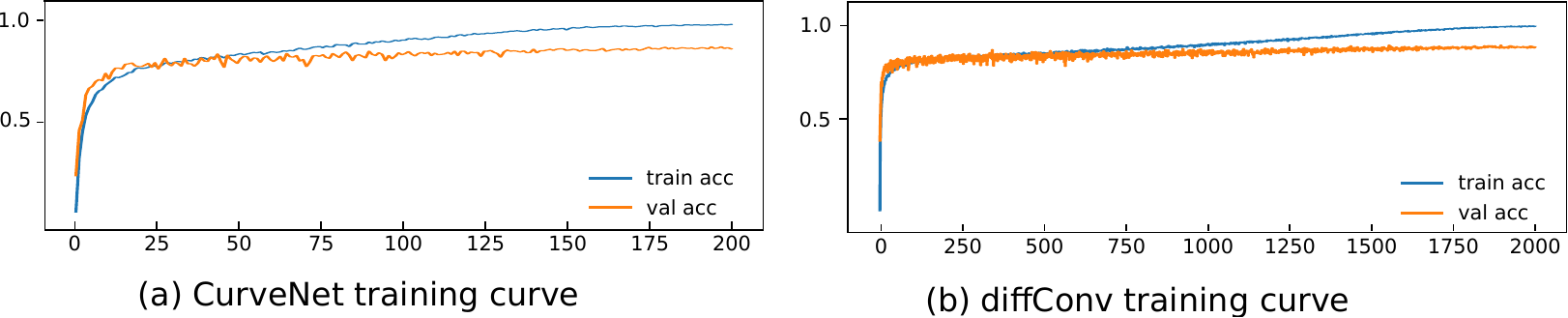}
  \caption{Training curves of our model and CurveNet in ModelNet40 classification task.}
  \label{fig:curve}
\end{figure}

\paragraph{Training process. }Our model is trained for more epochs for full convergence, compared to existing methods. Fig.~\ref{fig:curve} shows the learning curves of our model and CurveNet, and our training is as stable as CurveNet. The robustness study on ModelNet40-C also proves our model is not overfitting to the dataset. Note that this training setting does not affect our model's faster inference speed.

\section{Ablation Studies}
\paragraph{Ablation study over components.}
To verify the effectiveness of the proposed diffConv, we conducted a detailed ablation study over components on the ModelNet40 official split. The experimental settings and data augmentation was the same with the ModelNet40 classification task.

\setlength{\tabcolsep}{4pt}
\begin{table}
\begin{center}
  \caption{Ablation study over components. "LS" refers to Laplacian smoothing, "MAT" denotes masked attention, "DDBQ" stands for density-dilated ball query, and "BR" denotes balanced renormalization. Note that the equations refer to the equations in the main paper. }
  \label{tab:componentstudy}
\begin{tabular}{llllll}
\hline\noalign{\smallskip}
    LS  & DDBQ & MAT & BR & OA(\%)& MA(\%)\\
\noalign{\smallskip}
\hline
\noalign{\smallskip}
Eq. 3 & Eq. 8 & Eq. 9 & Eq. 11\\
\hline
\noalign{\smallskip}
     & & & & 89.8 & 85.2\\
     \checkmark & & & & 90.8 & 86.4\\
      & \checkmark & & & 90.6 & 86.1\\
      & & \checkmark & \checkmark & 92.7 & 89.9 \\
      \checkmark& \checkmark & & & 92.2 & 89.5\\
    \checkmark   & &\checkmark& \checkmark & 93.1 & 90.4\\
      & \checkmark &\checkmark & \checkmark & 92.5 & 89.2\\
    \checkmark &\checkmark  &\checkmark &  & 92.9 & 89.8\\  
     \checkmark &\checkmark  &\checkmark & \checkmark & \textbf{93.6} & \textbf{90.6}\\  
\hline

\end{tabular}
\end{center}
\end{table}
\setlength{\tabcolsep}{1.4pt}

We assessed the effect of employing the three main components of diffConv, namely Laplacian smoothing, density-dilated ball query, and masked attention.  
When Laplacian smoothing was not applied, the feature vector $S$ in Eq. 3 in the main paper was represented as $S=\hat{A}X$. When the density-dilated ball query was removed, the model grouped points by vanilla ball query with of a radius of $\sqrt{0.005}$. When masked attention was disabled, the adjacency matrix was replaced with the binary matrix similar to Eq. 5 in the main paper. We also evaluated the effect of the balanced renormalization strategy applied in masked attention. This ablation study was done by simply replacing the balanced renormalization with the normalization strategy employed by the original self-attention~\cite{vaswani2017attention}.

Table~\ref{tab:componentstudy} reports the overall and mean-class accuracy under different component combinations. When the model is not applying diffConv, the overall accuracy is only $89.8\%$.
We consistently see that the three components all bring improvements to the results. In contrast to the constant-radius ball query, our density-dilated modification improves the OA by $0.5$. This is in accordance with our analysis in Section 3.3 in the main paper that the long-range information flow is boosted by the dilated neighborhood. We also notice that masked attention plays a key role in diffConv. Comparing the fifth and the last row, masked attention improves the results by $1.4$ and $1.1$ in OA and MA separately. Besides, the employed balanced renormalization strategy shows significant improvement on model performance, compared to the normalization adopted by the original self-attention.    

\paragraph{Component contribution to robustness. }
Our model shows strong robustness on the ModelNet40-C benchmark. We studied the contribution of the proposed density-dilated ball query and masked attention to the model corruption robustness. Specifically, when masked attention was disabled, the adjacency matrix was replaced
with the binary matrix as in the ablation study over components. When density-dilated ball query was disabled, we employed the KNN grouping, which is sensitive to noise points according to our analysis in Section 3.3 in the main paper. The models were trained on the ModelNet40 dataset and evaluated on the ModelNet40 benchmark by OA and MA, and the ModelNet40-C benchmark by the corruption error rate. Table~\ref{tab:rob_contri} illustrates the experiment results. 

\setlength{\tabcolsep}{4pt}
\begin{table}
\begin{center}
  \caption{Component contribution to robustness. Here, "w/o DDBQ" denotes the model grouping 20 nearest neighbors instead of density-dilated ball query; "w/o MAT" stands for the model without masked attention; "Complete model" denotes the model equipped with all the proposed components. }
  \label{tab:rob_contri}
\begin{tabular}{lllll}
\hline\noalign{\smallskip}
    Model types & CER(\%) & OA(\%) & MA(\%)\\
\noalign{\smallskip}
\hline
\noalign{\smallskip}
    w/o DDBQ & 25.2 & 92.7 & 89.2 \\
    w/o MAT & 21.9 & 92.2 & 89.5 \\
\hline
\noalign{\smallskip}
    Complete model & \bf 21.4 & \bf 93.6 & \bf 90.6\\
\hline
\end{tabular}
\end{center}
\end{table}
\setlength{\tabcolsep}{1.4pt}

The results show that both the proposed density-dilated ball query and masked attention, especially the irregular ball query, contribute to the model's robustness. This supports our hypothesis of the robustness of the irregular point representation.    

\paragraph{Attention v.s.~inductive bias.}
According to Table~\ref{tab:componentstudy}, masked attention, which assigns each neighbor a weight, is the most effective part of diffConv. The weight does not rely on an inductive bias and is purely learned from point features as well as coordinates and updated dynamically during training. This is different from the predefined rules for point weighting applied in previous work. We compare masked attention with several intuitive inductive biases in Table~\ref{tab:bias}. The models were trained and evaluated on the ModelNet40 official split. The isotropic bias~\cite{qi2017pointnet++} treats all the neighbors equally. Spatial distance~\cite{yi2017syncspeccnn} and feature distance bias~\cite{xu2020learning} assign larger importance to the neighbor closer to the key point in the Euclidean and feature space respectively. The inverse density bias is taken from PointConv \cite{wu2019pointconv}, which posits high-density neighbors a lower contribution.
We implemented the last three biases via replacing the adjacency matrix from Eq. 9 in the main paper with the respective metrics processed by a Gaussian kernel, similar to~\cite{xu2020learning}.  

\setlength{\tabcolsep}{4pt}
\begin{table}
\begin{center}
  \caption{Results of study over attention v.s.~inductive bias. }
  \label{tab:bias}
\begin{tabular}{llll}
\hline\noalign{\smallskip}
    Aggregation rules & OA(\%) & MA(\%) \\
\noalign{\smallskip}
\hline
\noalign{\smallskip}
    Isotropic bias & $92.2$ ($1.4\downarrow$) & $89.5$ ($1.1\downarrow$)\\
    Spatial distance & $90.2$ ($3.4\downarrow$) & $84.9$ ($5.7\downarrow$)\\
    Feature distance & $91.1$ ($2.5\downarrow$) & $86.8$ ($3.8\downarrow$)\\
    Inverse density & $89.7$ ($3.9\downarrow$) & $85.0$ ($5.6\downarrow$)\\
\hline
\noalign{\smallskip}
    Ours & \textbf{93.6} & \textbf{90.6}\\
\hline
\end{tabular}
\end{center}
\end{table}
\setlength{\tabcolsep}{1.4pt}

Our method outperforms all the conventional inductive biases by more than $1.4$ and $1.1$ in OA and MA. With the introduction of pre-defined neighbor preference (rows 2, 3, and 4), the model performance becomes even worse than the isotropic bias (row 1) that treats all the neighbors evenly. In contrast to prior knowledge, the irregularity given by the density-dilated view and masked attention better exploits latent point local structure. We attribute the improvement to the introduction of irregularity.

\paragraph{Impact of bandwidth in kernel density estimation. }
Table~\ref{tab:h} presents the impact of bandwidth $h$ in kernel density estimation (Eq. 7 in the main paper) on ModelNet40 classification. According to the table, 0.1 is the optimal bandwidth. 

\setlength{\tabcolsep}{4pt}
\begin{table}
\begin{center}
  \caption{Results of our model with different kernel density bandwidths. }
  \label{tab:h}
\begin{tabular}{llll}
\hline\noalign{\smallskip}
    Kernel density bandwidth ($h$) & OA(\%) & MA(\%) \\
\noalign{\smallskip}
\hline
\noalign{\smallskip}
    0.05 & 93.1 & 90.1 \\
    \bf 0.1 & \bf 93.6 & \bf 90.6 \\
    0.5 & 93.3 & 90.1 \\
\hline
\end{tabular}
\end{center}
\end{table}
\setlength{\tabcolsep}{1.4pt}

\paragraph{Impact of squared initial searching radius. }
The impact of various squared initial searching radius $r^2$ settings is illustrated in Table~\ref{tab:r2}. According to Eq. 8 in the main paper, $r^2$ determines the lower bound of the searching radius. All the experiments were conducted on ModelNet40.

\setlength{\tabcolsep}{4pt}
\begin{table}
\begin{center}
  \caption{Results of our model with different squared initial searching radius. }
  \label{tab:r2}
\begin{tabular}{llll}
\hline\noalign{\smallskip}
     Squared initial searching radius ($r^2$) & OA(\%) & MA(\%) \\
\noalign{\smallskip}
\hline
\noalign{\smallskip}
    0.001 & 92.9 & 89.8 \\
    \bf 0.005 & \bf 93.6 & \bf 90.6 \\
    0.01 & 93.2 & 90.4 \\
    0.05 & 93.0 & 90.1 \\  
    0.1 & 91.8 & 88.1 \\  
\hline
\end{tabular}
\end{center}
\end{table}
\setlength{\tabcolsep}{1.4pt}

In line with the results, we find that with a large $r^2$, the model performance degenerates, since the model fails to capture point local geometric structures. 

\paragraph{Ablation study over dilating strategies. }We compared three different strategies for dilating searching radius in Eq. 8 in the main paper. Given $r$ the pre-set initial searching radius, in strategy one, the dilated radius $r_i=r(1+\hat{d}_i)$ is linearly correlated with the point kernel density $\hat{d}_i$. In strategy two, the squared dilated radius $r_i^2=r^2(1+\hat{d}_i)$ is linearly correlated with the point kernel density. In the last strategy, the dilated radius $r_i=r\cdot (1+\frac{e^{\hat{d}_i}-1}{e-1})$ has a nonlinear relationship with the kernel density. All the experiments were run on the ModelNet40 benchmark.

\setlength{\tabcolsep}{4pt}
\begin{table}
\begin{center}
  \caption{Results of study over different radius dilating strategies. "Linear-1", "Linear-2" and "Exponent" denote the three strategies in the text respectively. }
  \label{tab:ddbqc}
\begin{tabular}{llll}
\hline\noalign{\smallskip}
     Strategies & OA(\%) & MA(\%) \\
\noalign{\smallskip}
\hline
\noalign{\smallskip}
    Linear-1 & 92.6 & 89.6 \\
    \bf Linear-2 & \bf 93.6 & \bf 90.6 \\
    Exponent & 93.0 & 89.9 \\
\hline
\end{tabular}
\end{center}
\end{table}
\setlength{\tabcolsep}{1.4pt}

Table~\ref{tab:ddbqc} presents the results. The second strategy, "Linear-2", achieves the best performance. This demonstrates that a too-fast dilation speed fails to benefit the point feature learning. 

\section{Additional Visualizations of Masked Attention}
\begin{figure}
\centering
\includegraphics[width=\linewidth]{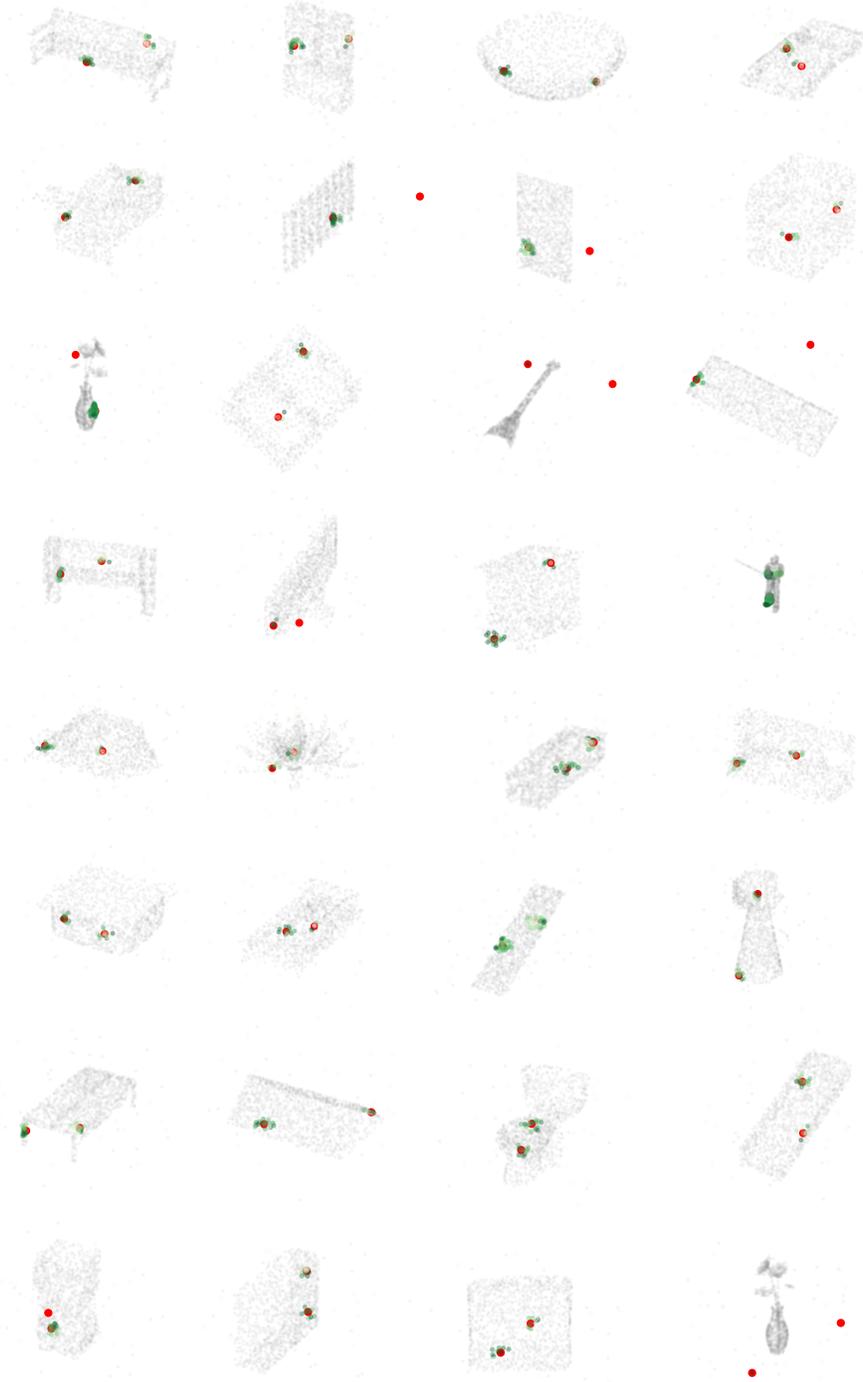}
\caption{Masked attention maps for ModelNet40-C. Red dots are picked key points; green dots denote neighbors; the color scale denotes the masked attention score. Zoom in for details. }
\label{fig:att_extra_vis}
\end{figure}

Fig. 4 in the main paper illustrates the attention maps of 8 objects from different categories on the ModelNet40-C benchmark. We visualize the attention maps of objects from the rest 32 categories in   Fig.~\ref{fig:att_extra_vis}. The attention scores were taken from the second diffConv of the classification network. All the objects were corrupted by "the most severe background noise" (a "severity" of 5). For each object, we randomly picked two key points and visualized their neighbors. As shown in the figure, our diffConv isolates the noise points, endows the flat-area points with a larger receptive field and focuses on the neighbors with larger differences in geometric features to the key points. 

\begin{figure}[h]
  \centering
  \includegraphics[width=\linewidth]{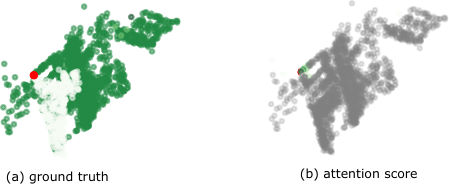}
  \caption{Ground truth mask and attention score of an example from Toronto3D. Zoom in for details. }
  \label{fig:att}
\end{figure}

We also illustrate how neighbors are selected when two flat surface approach each other, with the ground truth mask and attention score of an example point cloud from Toronto3D in Fig.~\ref{fig:att}. The red point is a building point that lies on the boundary of the building (green points) and the road (milky points). The building and the road are approximately flat surfaces. According to the figure, our diffConv emphasizes the neighbors from the building (with darker color). 

\end{document}